\pdfoutput=1

\documentclass[11pt]{article}

\usepackage{acl}

\usepackage{times}
\usepackage{latexsym}

\usepackage[T1]{fontenc}

\usepackage[utf8]{inputenc}

\usepackage{microtype}

\usepackage{color}
\usepackage{soul}
\usepackage{amsmath}
\usepackage{amsfonts}
\usepackage{xcolor}
\usepackage{graphicx}
\usepackage[shortlabels]{enumitem}
\usepackage{booktabs}
\usepackage{multirow}
\usepackage{array}
\usepackage{float}
\usepackage{arydshln}
\usepackage{adjustbox}
\usepackage{multicol}
\usepackage{mathabx}
\usepackage[section]{placeins}
\usepackage[normalem]{ulem}

\usepackage[multiple]{footmisc}

\title{Neural Pipeline for Zero-Shot Data-to-Text Generation}

\author{Zdeněk Kasner \and Ondřej Dušek \\
  Charles University, Faculty of Mathematics and Physics\\
  Institute of Formal and Applied Linguistics \\
  Prague, Czech Republic \\
  \texttt{\{kasner,odusek\}@ufal.mff.cuni.cz}
}

\usepackage{xcolor}
\usepackage[normalem]{ulem}

\definecolor{lightblue}{RGB}{4, 89, 186}
\newcommand{\lightblue}[1]{{\leavevmode\color{lightblue}{#1}}}
\newcommand{\green}[1]{{\leavevmode\color{green!70!black!100}{#1}}}
\newcommand{\red}[1]{{\leavevmode\color{red!70!black!100}{#1}}}
\newcommand{\orange}[1]{{\leavevmode\color{orange}{#1}}}

\newcommand{\baselinecopy}{\textsc{copy}}

\newcommand\Tstrut{\rule{0pt}{2.2ex}}
\newcommand\Bstrut{\rule[-0.6ex]{0pt}{0pt}}

\begin{document}
\maketitle

\begin{abstract}
In data-to-text (D2T) generation, training on in-domain data leads to overfitting to the data representation and repeating training data noise. We examine how to avoid finetuning  pretrained language models (PLMs) on D2T generation datasets while still taking advantage of surface realization capabilities of PLMs. Inspired by pipeline approaches, we propose to generate text by transforming single-item descriptions with a sequence of modules trained on general-domain text-based operations: ordering, aggregation, and paragraph compression. We train PLMs for performing these operations on a synthetic corpus \textsc{WikiFluent} which we build from English Wikipedia. Our experiments on two major triple-to-text datasets---WebNLG and E2E---show that our approach enables D2T generation from RDF triples in zero-shot settings.\footnote{Our code and data is available at \url{https://github.com/kasnerz/zeroshot-d2t-pipeline}.\label{fn:url}}
\end{abstract}

\section{Introduction}
The aim of data-to-text (D2T) generation is to produce natural language descriptions of structured data \cite{gatt2018survey,reiter1997building}. Although pipelines of rule-based D2T generation modules are still used in practice \cite{dale2020natural}, end-to-end approaches based on PLMs recently showed superior benchmark performance \cite{ke2021jointgt,chen-etal-2020-kgpt,ferreira20202020, kale-rastogi-2020-text, ribeiro2021investigating}, surpassing pipeline systems \cite{ferreira2019neural} in both automatic and human evaluation metrics.

\begin{figure}[t]
  \centering
  \includegraphics[width=\columnwidth]{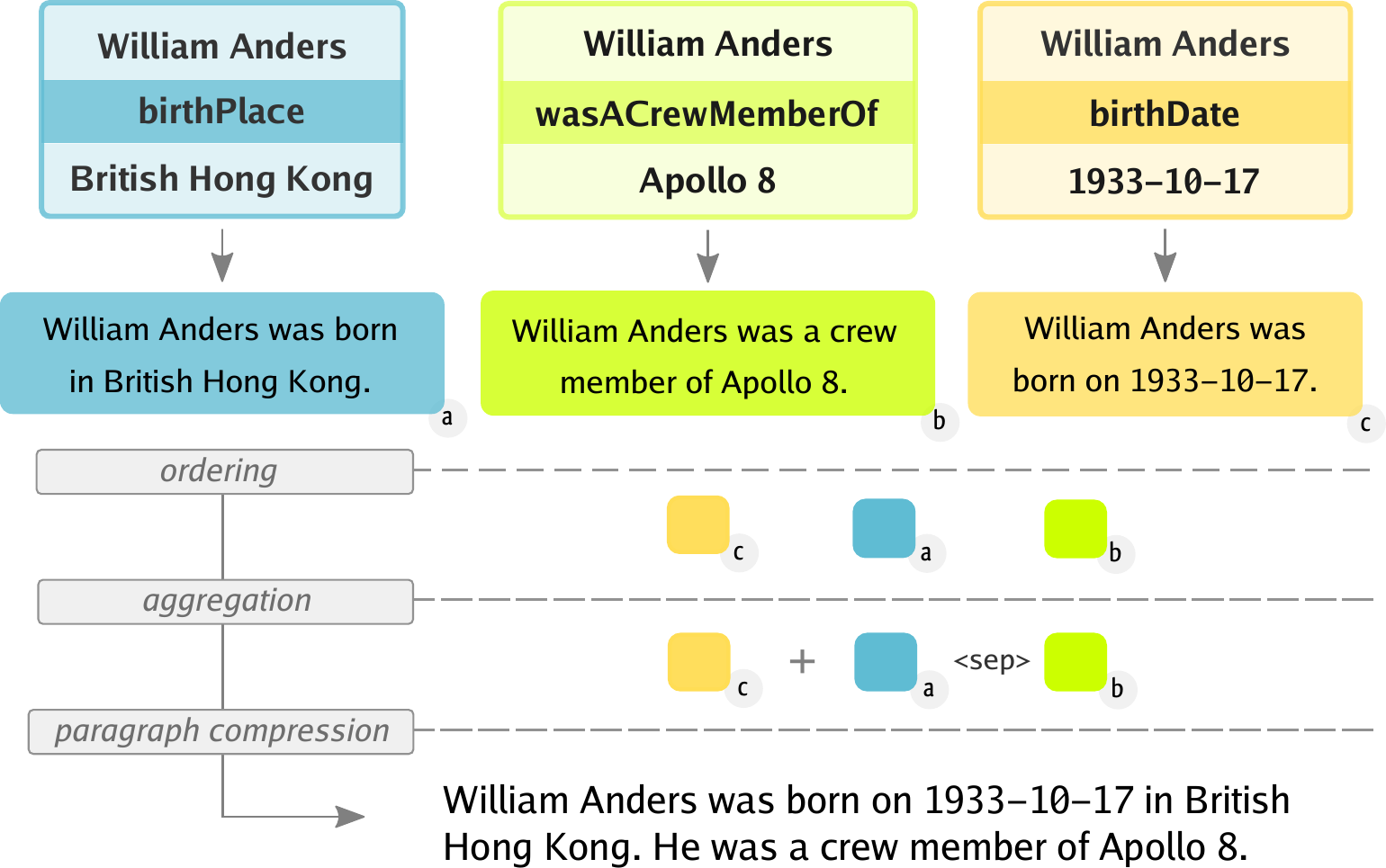}
  \caption{A scheme of our pipeline for zero-shot data-to-text generation from RDF triples: (1) ordering, (2) aggregation, (3) paragraph compression. Individual pipeline modules are trained on a large general-domain text corpus and operate over text in natural language. In-domain knowledge is included only in the simple hand-crafted templates for each predicate.}\label{fig:fnr}
\end{figure}

Finetuning PLMs on human-written references is widely accepted as a standard approach for adapting PLMs to the D2T generation objective and achieving good performance on a given benchmark \cite{agarwal2021knowledge,ke2021jointgt}. However, finetuning a model on the domain-specific data leads to overfitting to the particular benchmark, decreasing performance on out-of-domain data \cite{laha2020scalable}. Gathering a large set of references for a particular domain is also costly and time-consuming as it usually requires collecting human-written references through crowdsourcing \cite{duvsek2020evaluating_challenge}. These problems can be partially mitigated using \emph{few-shot} approaches \cite{chen-etal-2020-shot,ke2021jointgt,su2021few}, which operate with only several dozens or hundreds of annotated examples, but the robustness of these approaches is questionable---selecting a representative set of examples which would improve performance is difficult \cite{chang2021training}, and the limited sample is often noisy, increasing the chance of hallucinations and omissions \cite{duvsek2019semantic,harkous2020have,rebuffel2021controlling}.

In this paper, we present a \emph{zero-shot} alternative to the traditional finetuning paradigm by formulating the D2T generation from RDF triples as a sequence of general-domain operations over text in natural language. We start by transforming individual triples to text using trivial templates, which we subsequently order, aggregate, and compress on the paragraph level to produce the resulting description of the data. In constrast to traditional pipeline systems, all our pipeline modules are built upon PLMs and operate over sentences in natural language. The modules are trained on our new \textsc{WikiFluent} corpus, which contains 934k examples of first paragraphs from the English Wikipedia, each supplied with a synthesized set of simple template-like sentences which together convey the meaning of the original paragraph. 
Our approach allows generating natural language descriptions from RDF triples with a minimum amount of domain-specific rules or knowledge and without using training data from the D2T datasets. Although our approach is primarily a probe into the territory of zero-shot approaches and cannot yet match the quality of state-of-the-art models, we show that it can yield large improvements upon simple baselines and match older supervised systems on automatic metrics for text fluency. Moreover, the semantic accuracy metrics and our manual error analysis suggest that our approach offers a way to prevent omissions and hallucinations common in few-shot approaches.

Our contributions are the following:
\begin{enumerate}[label={(\arabic*)},nosep,leftmargin=17pt]
  \item We propose an alternative D2T generation approach based on general-domain text-to-text operations (ordering, aggregation, and paragraph compression).
  \item We introduce a synthetic \textsc{WikiFluent} corpus containing 934k sentences based on English Wikipedia, providing training data for the operations in (1).
  \item We apply our system on two D2T datasets and evaluate its performance both automatically and manually, including the contribution of individual pipeline modules.
  \item We release our code, data, pretrained models, and system outputs to ease future research.\textsuperscript{\ref{fn:url}}
\end{enumerate}

\section{Related Work}
\label{sec:related}
\paragraph{D2T Generation with PLMs} Large neural language models pretrained on self-supervised tasks \cite{lewis2020bart,liu2019roberta,devlin2019bert} have recently gained a lot of traction in D2T generation research \cite{ferreira20202020,kasner_train_2020}. Following \citet{chen-etal-2020-shot}, other works adopted PLMs for few-shot D2T generation \cite{chang2021neural,su2021few}. \citet{kale-rastogi-2020-text} and \citet{ribeiro2021investigating} showed that PLMs using linearized representations of data can outperform graph neural networks on graph-to-text datasets, recently surpassed again by graph-based models \cite{ke2021jointgt,chen-etal-2020-kgpt}. Although the models make use of general-domain pretraining tasks, all of them are eventually finetuned on domain-specific data.

\paragraph{Pipeline-based D2T Generation} Until the recent surge of end-to-end approaches \cite{duvsek2020evaluating_challenge}, using several modules connected in a pipeline was a major approach for D2T generation \cite{gatt2018survey,reiter2007architecture,reiter1997building}. Our approach is inspired by the pipeline approaches, in particular the pipelines utilizing neural modules \cite{ferreira2019neural}. In contrast with these approaches, our pipeline works with unstructured data in natural language and it operates in zero-shot setting, i.e.\ without using any training data from target D2T datasets.

\citet{laha2020scalable} introduce a three-step pipeline for zero-shot D2T generation similar to ours. Unlike the approach we describe here, they use a semi-automatic template generation system,\footnote{As we describe in §\ref{sec:templates_model}, we opted for a simpler way for generating the templates to showcase the results of our approach independently of the template generator quality.} their sentence fusion is rule-based, and they do not address content planning.

\paragraph{Content Planning in D2T Generation} Content planning, i.e. the task of ordering input facts and aggregating them into individual sentences, is one of the steps of the traditional D2T pipeline \cite{gatt2018survey}. As shown by \citet{moryossef2019improving,moryossef2019step} and confirmed by other works \cite{puduppully2019data,zhao2020bridging,trisedya2020sentence,su2021plan}, including a content plan improves the quality of outputs in neural D2T pipelines. Unlike the aforementioned planners, which use predicates or keys from D2T datasets for representing the data items, our planner is trained on ordering sentences in natural language.

\paragraph{Sentence Ordering} Sentence ordering is the task of organizing a set of natural language sentences to increase the coherence of a text \cite{barzilay2001sentence,lapata2003probabilistic}. Several neural methods for this task were proposed, using either interactions between pairs of sentences \cite{chen2016neural,li2017neural}, global interactions \cite{gong2016end,wang2019hierarchical}, or combination of both \cite{cui2020bert}. We base our ordering module (§\ref{sec:ord_model}) on the recent work of \citet{calizzano2021ordering}, who use a pointer network \cite{wang2019hierarchical,vinyals2015pointer} on top of a PLM.

\paragraph{Aggregating Input into Sentences} Typically, multiple pieces of input information need to be merged into a single sentence. Previous works \cite{wiseman2018learning,shao-etal-2019-long,shen-etal-2020-neural,xu2021agggen} capture the segments which correspond to individual parts of the input as latent variables. Unlike these works, we adopt a simpler scenario using an already ordered sequence of facts (see §\ref{sec:templates}), into which we selectively insert delimiters to mark sentence boundaries.

\paragraph{Paragraph Compression} We introduce \textit{paragraph compression} (PC) as a new task and the final step in our D2T generation pipeline. This task combines several standard natural-language tasks including sentence fusion, rephrasing, and coreference resolution. Unlike text summarization or simplification \cite{zhang2020pegasus,jiang2020neural}, we aim to convey the complete semantics of the text without omitting any facts. In contrast to sentence fusion \cite{geva2019discofuse,barzilay2005sentence} or sentence compression \cite{filippova2013overcoming}, we operate in the context of multiple sentences in a paragraph.  The task is the central focus of our \textsc{WikiFluent} corpus (§\ref{sec:wikifluent}).

\section{Method}
\label{sec:method}
In this section, we provide the formal description of our proposed approach. We focus on the task of producing a natural language description $Y$ for a set of $n$ RDF triples $X = \{x_1, \ldots, x_n\}$. Each triple $x_i = \{s_i, p_i, o_i\}$ consists of subject $s_i$, predicate $p_i$, and object $o_i$. 

Our pipeline proceeds as follows. Given a set of triples $X$ on the input, we:
\begin{enumerate}[label={(\arabic*)},nosep,leftmargin=17pt]
  \item transform the triples into \textit{facts}, which are sentences in natural language,
  \item sort the facts using an \textit{ordering} module,
  \item insert sentence delimiters between the sorted facts using an \textit{aggregation} module,
  \item input the ordered sequence of facts with delimiters into a \textit{paragraph compression} module, which generates the final description $Y$.
\end{enumerate}

The individual steps are described in the following sections: transforming individual triples to text (§\ref{sec:templates}), ordering (§\ref{sec:ordering}), aggregation (§\ref{sec:agg}), and paragraph compression (§\ref{sec:pc}).

\subsection{Transforming Triples to Facts}
\label{sec:templates}
The first step in our pipeline involves transforming each of the input triples $x_i \in X$ into a fact $f_i \in F$  using a transformation $T: X \rightarrow F$. We define a fact $f_i$ as a single sentence in natural language describing $x_i$. 
The transformation serves two purposes: (a) preparing the data for the subsequent text-to-text operations, (b) introducing in-domain knowledge about the semantics of individual predicates. This step can be realized e.g. using a simple template for each predicate (cf. §\ref{sec:templates_model}).

\subsection{Ordering the Facts}
\label{sec:ordering}
We assume that the default order of triples $X$ is random and the same applies for the respective facts $F$. Note, however, that that $F$ is a indeed set of meaningful sentences. We can use this to our advantage and apply a sentence ordering model to maximize the coherency of the paragraph resulting from their concatenation. An example outcome of such operation may be grouping together facts mentioning \textit{birth date} and \textit{birth place} of a person, followed by their \textit{occupation} (see Figure \ref{fig:fnr}). The ordering module allows downstream modules to only focus on operations over neighboring sentences.

Formally, we apply the ordering model $O(F)$ to get an ordered sequence of facts: $F_o = \{f_{o_1}, \ldots, f_{o_n}\}$, where $o_{1:n}$ is a permutation of indices. We describe our ordering model in §\ref{sec:ord_model}.

\subsection{Aggregating the Facts}
\label{sec:agg}
Some facts will be typically mentioned together in a single sentence. Considering the previous example, \textit{occupation} is likely to be mentioned separately, while \textit{birth date} and \textit{birth place} are likely to be mentioned together. Using an ordered sequence of facts as input, we can apply an aggregation model to decide which facts should be merged into a single sentence. 

Formally, the aggregation model takes a sequence of ordered facts $F_o$ as input and produces a sequence of sentence delimiters $A(F_o) = \{\delta_{o_1}, \delta_{o_2}, \ldots, \delta_{o_{n-1}}\}$; $\delta_{i} \in \{0, 1\}$. The output $\delta_{i}=1$ means that the neighboring facts should be mentioned separately, i.e. the neighboring sentences should \textit{not} be fused. Conversely, $\delta_{i}=0$ means that the facts should be aggregated and their corresponding sentences should be fused. We describe our aggregation model in §\ref{sec:agg_model}.

\subsection{Paragraph Compression}
\label{sec:pc}
The paragraph compression (PC) model is a generative model which outputs the final text description.  It has two main objectives: (a) \textit{fusing} related sentences, i.e., sentences $i$ and $j$ in between which $\delta_{i}=0$, and (b) \textit{rephrasing} the text to improve its fluency, e.g. fixing disfluencies in the templates, replacing noun phrases with refering expressions, etc. The goal of the task is to preserve the semantics of the text which is an already ordered sequence of sentences, so the edits will typically be minor.  Formally, the model takes as input the ordered sequence of facts with delimiters $F_a = \{f_{o_1}, \delta_{o_1}, f_{o_2}, \ldots, \delta_{o_{n-1}}, f_{o_n}\}$ and produces the final text $Y$.  We describe our PC model in §\ref{sec:pc_model}.

\section{\textsc{WikiFluent} Corpus}
\label{sec:wikifluent}
Here we descibe the process of building a large-scale synthetic corpus \textsc{WikiFluent}. The corpus provides training data for the neural models which we use in our implementation of the ordering, aggregation, and paragraph compression modules (cf.~§\ref{sec:implementation}).

Our goal is to cover a broad range of domains while capturing the sentence style in D2T generation with respect to both the input facts and the target descriptions. In other words, we aim to build a corpus in which (1) the input is a set of simple, template-like sentences, (2) the output is a fluent text in natural language preserving the semantics of the input. As we describe below in detail, we achieve that by using human-written paragraphs in English Wikipedia and applying \textit{split-and-rephrase} and  \textit{coreference resolution} models to obtain synthetic source texts. The process is illustrated in Figure \ref{fig:wikifluent}; corpus statistics are included in Appendix \ref{app:stats}.

\subsection{Data Source} For building the \textsc{WikiFluent} corpus, we extracted 934k first paragraphs of articles from a Wikipedia dump\footnote{\texttt{enwiki-20210401-pages-articles-multistream}} using WikiExtractor \cite{Wikiextractor2015}. Wikipedia is commonly used for large-scale pretraining of D2T generation models \cite{jin2020genwiki,chen-etal-2020-kgpt}. Although it is not bias-free, it provides more balanced sample of natural language use than typical D2T generation datasets. We used the first paragraphs of Wikipedia entries, which contain mostly concise, fact-based descriptions. 
We selected paragraphs with length between 30-430 characters; filtering out lists, disambiguations, and repeated and malformed paragraphs. To balance the length of inputs, we selected 250k examples each from 4 equally sized length ranges (30-130 characters, etc.).

\begin{figure}[t]
  \centering
  \includegraphics[width=\columnwidth]{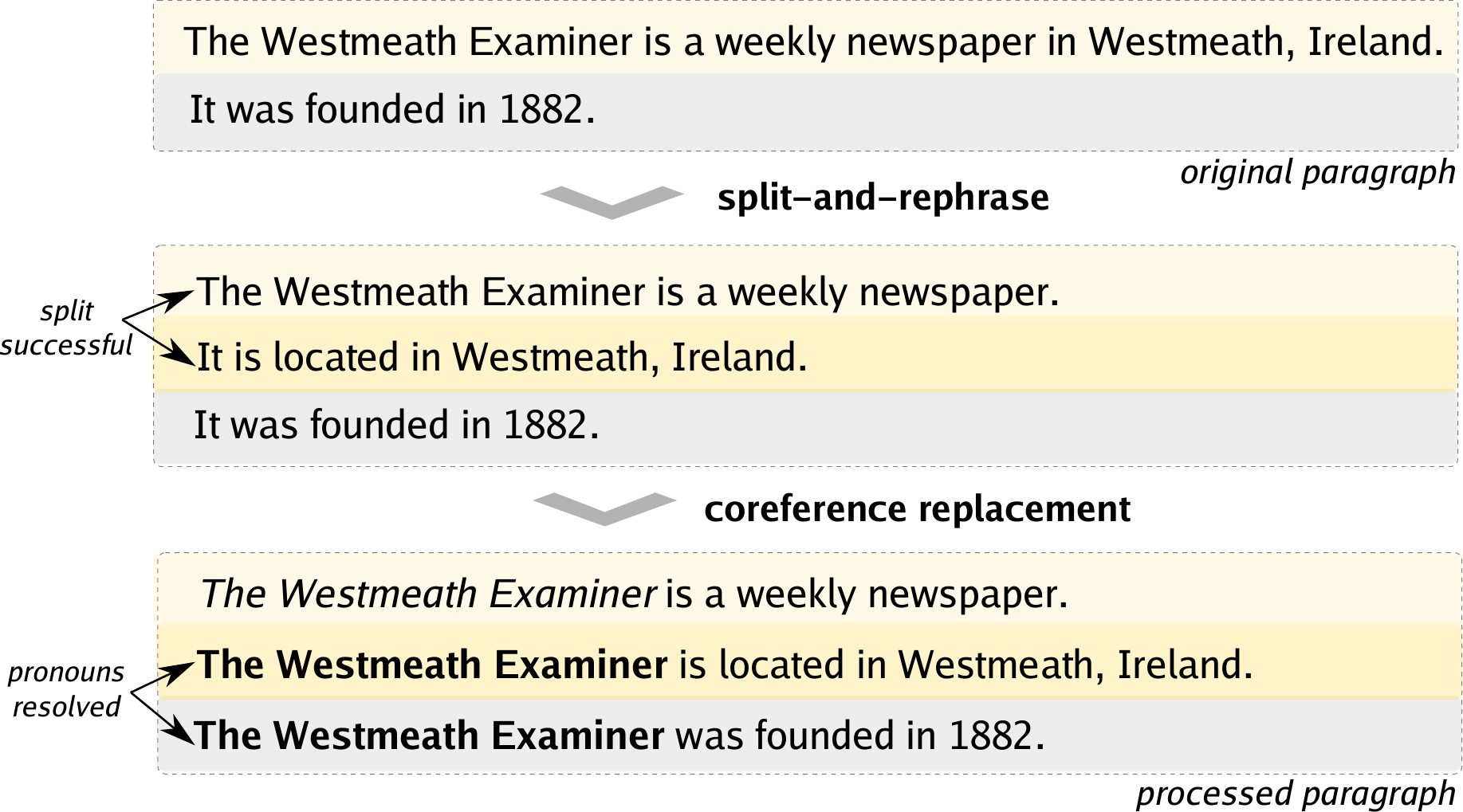}
  \caption{The building process of the \textsc{WikiFluent} corpus. We apply a split-and-rephrase model on each sentence in the paragraph and resolve coreferences in the split sentences. The result is a set of simple sentences which together convey the same meaning as the original paragraph. The synthesized sentences are used as \textit{input} into our models, the original human-written texts are used as \textit{ground truth}.}\label{fig:wikifluent}
\end{figure}

\subsection{Split-and-Rephrase} 

To generate a set of simple sentences, we divide each paragraph into sentences using NLTK \cite{bird2006nltk} and apply a \textit{split-and-rephrase} model on each sentence. Split-and-rephrase is a task of splitting a complex sentence into a meaning preserving sequence of shorter
sentences \citep{narayan-etal-2017-split}. The process is illustrated in the upper part of Figure \ref{fig:wikifluent}. 

We train our split-and-rephrase model on the large-scale WikiSplit corpus by \citet{botha-etal-2018-learning}, containing human-made sentence splits from Wikipedia edit history. Following the same setup as for a paragraph compression model (§\ref{sec:pc}), we train BART-base \cite{lewis2020bart} on the WikiSplit dataset in a sequence-to-sequence setting. Next, we apply the trained split-and-rephrase model on each sentence in our Wikipedia-based corpus, uniformly randomly choosing between 0-2 recursive calls to ensure that the splits are not deterministic. If the sentence cannot be meaningfully split, the model tends to duplicate the sentence on the output; in that case, we use only the original sentence and do not proceed with the splitting.

\subsection{Coreference Replacement}  As the next step, we concatenate the split sentences and apply a coreference resolution model \cite{gardner2018allennlp,Lee2018HigherorderCR} in order to replace referring expressions with their antencendents (e.g., pronouns with noun phrases). The motivation for this step is to match the style of the facts (see §\ref{sec:templates}), which do not use pronouns since each fact describes a single triple only. Note that this procedure replaces the referring expressions only in the synthesized sentences (which are used as input) and keeps them in the original paragraphs (which are used as ground truth). As a consequence, the paragraph compression module is implicitly trained to generate referring expressions in the final description.

\subsection{Filtering} To ensure that the generated sentences convey the same semantics as the original paragraph, we use a pretrained RoBERTa model\footnote{\url{https://huggingface.co/roberta-large-mnli}} \cite{liu2019roberta} trained on the MultiNLI dataset \cite{williams2018broad} for checking the semantic accuracy of the generated text. Following \citet{duvsek2020evaluating}, we test if the original paragraph entails each of the synthesized sentences (checking for omissions), and if the set of concatenated synthesized sentences entails the original paragraph (checking for hallucinations). In a filtered version of the \textsc{WikiFluent} corpus, we include only the examples without omissions or hallucinations (as computed by the model), reducing it to 714k examples (approximately 75\% of the original size).

\section{Implementation}
\label{sec:implementation}

In this section, we describe how we implement our pipeline modules (§\ref{sec:method}) using simple template transformations (§\ref{sec:templates_model}) and neural models trained on the \textsc{WikiFluent} dataset (§\ref{sec:ord_model}-\ref{sec:pc_model}).\footnote{Our training setup details are included in Appendix \ref{app:setup}.}

\begin{figure*}[t]
  \centering 
  \includegraphics[width=\textwidth]{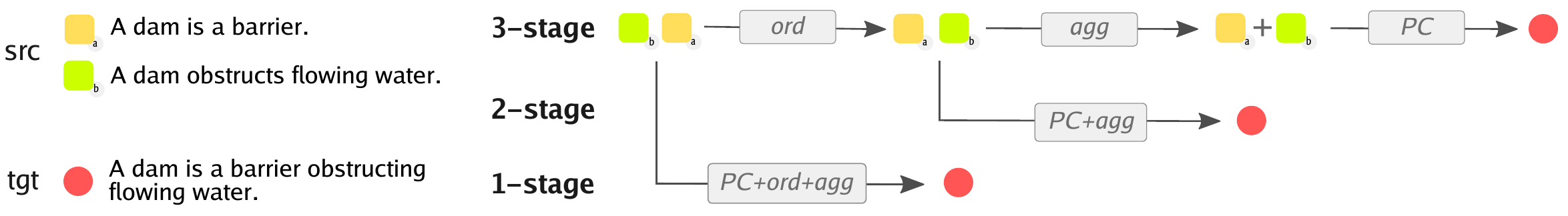}

  \caption{An example illustrating how the individual modules are trained and subsequently applied as the parts of the pipeline. See §\ref{sec:ord_model} for description of the ordering model (\textsc{ord}), §\ref{sec:agg_model} for the aggregation model (\textsc{agg}), and §\ref{sec:pc_model} and §\ref{sec:experiments} for the paragraph compression model (\textsc{PC, PC+agg, PC+ord+agg}).}\label{fig:pipeline}
\end{figure*}

\subsection{Templates}
\label{sec:templates_model}

\begin{table}[t]\centering\footnotesize
  \begin{tabular}{lll} \toprule
     \textbf{dataset} & \textbf{predicate} & \textbf{template}  \\ \midrule
      \multirow{3}{*}{\textbf{WebNLG}} & instrument & \textit{<$s$> plays <$o$>.}  \\
      & countryOrigin & \textit{<$s$> comes from <$o$>.}\\
      & width & \textit{<$s$> is <$o$> wide.} \Bstrut \\\hdashline[0.5pt/2pt] 
      \multirow{3}{*}{\textbf{E2E}} & eatType & \textit{<$s$> is a <$o$>.}\Tstrut \\
      & food &\textit{<$s$> serves <$o$> food.}\\
      & area & \textit{<$s$> is in the <$o$>.} \\ \bottomrule
     \end{tabular}
  \caption{Examples of templates for predicates in the WebNLG and E2E datasets with placeholders for the subject (<$s$>) and the object (<$o$>).}\label{tab:templates}
  \end{table}

We transform triples into facts (§\ref{sec:templates}) using a single-triple template $t_i$ for each predicate. For example, if $p_i=\textit{instrument}$, then \mbox{$T(p_i)=\textit{``}s_i\textit{ plays }o_i\textit{''}$} (cf.~Table \ref{tab:templates}). We follow previous work in which simple hand-crafted templates have been used as an efficient way of introducing domain knowledge \cite{kale-rastogi-2020-template,kasner2020data}. Compared to more complex rule-based template generation engines \cite{laha2020scalable,heidari2021getting,mehta2021improving}, the approach may produce less fluent outputs, but it minimizes manual workload and makes it easier to control the quality of the input for the subsequent steps.

\subsection{Ordering Model} 
\label{sec:ord_model}
For our ordering model (§\ref{sec:ordering}), we use the \textit{Simple Pointer} model from \citet{calizzano2021ordering}. The model is based on a pretrained BART-base extended with a pointer network from \citet{wang2019hierarchical}. We provide a short description of the model here; for details please refer to \citet{calizzano2021ordering}.

In the encoding phase, facts $F$ are concatenated and tokenized. Each fact is surrounded by special tokens denoting the beginning (\texttt{<s>}) and the end (\texttt{</s>}) of the fact. The sequence is processed by the BART encoder, generating a sequence of encoder states $E$ for each end token \texttt{</s>} representing the preceding fact.

The decoding proceeds autoregressively. To bootstrap the decoding process, the pair of tokens \texttt{<s></s>} is fed into the decoder, producing the decoder state $d_1$. The pointer network (attending to $d_1$ and $E$), selects the first ordered fact $f_{o_1}$, which is fed into the decoder in the next step ($d_2 = $\texttt{<s>$f_{o_1}$</s>}). The process is repeated until the all the facts are decoded in a particular order.

The pointer network computes the probability of a fact to be on the $j$-th position, using the encoder output $E$ and the decoder output state $d_j$. The network is based on the scaled dot product attention, where $d_j$ is the query and encoder outputs $E_i$ are the keys:
\begin{gather*}
  Q = d_j W_Q \\
  K = E W_K \\
  P_{j} = \operatorname{softmax}\left(\frac{QK^T}{\sqrt{b}}\right).
\end{gather*}
Here $W_Q$ and $W_K \in \mathbb{R}^{b\times b}$, $b$ is the dimension of BART hidden states, and $P_{j} \in \mathbb{R}^{n+1}$ is the probability distribution for the $j$-th position (i.e., $P_{ji}$ is the probability that fact $f_i$ is on the $j$-th position).

We train the model using the synthesized simple sentences in the \textsc{WikiFluent} corpus, randomly shuffling the order of the sentences and training the model to restore their original order.

\subsection{Aggregation Model}
\label{sec:agg_model}
We base our aggregation model (§\ref{sec:agg}) on RoBERTa-large \cite{liu2019roberta} with a token classification head.\footnote{\url{https://huggingface.co/transformers/model\_doc/roberta.html\#robertafortokenclassification}} Similarly to the ordering model (§\ref{sec:ord_model}), we input the sequence of (now ordered) facts $F_o$ into the model, separating each pair of facts $f_{o_i}$ with a special token \texttt{</s>} (used by the model as a separator). Subsequently, the token classification layer classifies each separator $\texttt{</s>}_i$ position into two classes $\{0,1\}$ corresponding to the delimiter $\delta_i$. We ignore the outputs for the non-separator tokens while computing cross-entropy loss.

We create the training examples using the synthesized sentences in the \textsc{WikiFluent} corpus, in which we set $\delta_i=0$ for the sentences $i,i+1$ which were originally aggregated (i.e., are the result of splitting a single sentence) and $\delta_i=1$ otherwise.

\subsection{Paragraph Compression Model}
\label{sec:pc_model}
We adopt BART-base for our paragraph compression model. We finetune the model on the \textsc{WikiFluent} corpus, concatenating the synthesized sentences on the input. We add delimiters between the sentences $i$ and $i+1$ where $\delta_i=1$ using a special token \texttt{<sep>}, which we add to the model vocabulary. As shown in \citet{keskar2019ctrl}, including control codes for training the model can steer the model towards producing certain outputs. Here we expect that the model will learn to fuse the sentences between which there are no delimiters on the input. We evaluate how the model learns to respect the order and aggregation markers in §\ref{sec:eval_planning}.

\section{Experiments}
\label{sec:experiments}
We train our pipeline modules on the \textsc{WikiFluent} corpus as described in §\ref{sec:implementation}. Next, we use these modules \textit{without finetuning} for generating descriptions for RDF triples on two English D2T datasets, WebNLG and E2E. 

\paragraph{Datasets} The datasets differ in domain, size, textual style, and number of predicates (see Appendix \ref{app:stats} for details):
\begin{itemize}[nosep,leftmargin=17pt]
    \item \textbf{WebNLG} \cite{gardent2017webnlg,ferreira20202020} contains RDF triples from DBPedia \cite{auer2007dbpedia} and their crowdsourced descriptions. We use version 1.4 of the dataset for comparison to prior work. We hand-crafted templates for all 354 predicates, including unseen predicates in the test set.\footnote{See Appendix \ref{app:templates} for details on template creation.}
    \item\textbf{E2E} \cite{novikova_e2e_2017,duvsek2020evaluating_challenge} contains restaurant recommendations in the form of attribute-value pairs. We use the cleaned version of the dataset \cite{duvsek2019semantic}. Following previous work, we transform the attribute-value pairs into RDF triples (using the restaurant name as a subject) and then apply the same setup as for WebNLG. We created a template for each of the 8 attributes manually. 
\end{itemize}

\paragraph{Pipeline versions} In order to evaluate individual components of our pipeline, we train three versions of the \textit{paragraph compression} model  (see §\ref{sec:pc_model}). The models share the same architecture and targets, but differ in their inputs:
\begin{itemize}[nosep,leftmargin=17pt]
  \item \textsc{PC} -- the model takes as an input ordered facts with delimiters (as described in §\ref{sec:pc}),
  \item \textsc{PC+agg} -- the model takes as an input ordered facts \textit{without} delimiters (i.e., the aggregation is left implicitly to the model),
  \item \textsc{PC+ord+agg} -- the model takes as an input facts in \textit{random} order and \textit{without} delimiters (i.e., both ordering and aggregation are left implicitly to the model).
\end{itemize}

Correspondingly, we test three versions of the pipeline in our \textbf{ablation study} (see Figure~\ref{fig:pipeline}):
\begin{itemize}[nosep,leftmargin=17pt]
  \item \textsc{3-stage} -- a full version of the pipeline consisting of the ordering model (\textsc{ord}), the aggregation model (\textsc{agg}) and the \textsc{PC} model (following the full pipeline from §\ref{sec:method}),
  \item \textsc{2-stage} -- a pipeline consisting of the \textsc{ord} model and the \textsc{PC+agg} model,
  \item \textsc{1-stage} -- a single stage consisting of the \textsc{PC+ord+agg} model.
\end{itemize}

We evaluate all versions of the pipeline with PC models trained on the \textit{full} and \textit{filtered} versions of the \textsc{WikiFluent} dataset (see §\ref{sec:wikifluent}).

\section{Evaluation and Discussion}

\label{sec:eval_disc}
Our main aim is the evaluation of our pipeline on the downstream task of D2T generation.
We evaluate outputs from the \textsc{\{1,2,3\}-stage} variants of our pipeline using automatic metrics (§\ref{sec:eval_auto}), and we perform a detailed manual error analysis of the model outputs (§\ref{sec:eval_manual}). We also evaluate the performance of the content planning modules and the ability of the PC module to follow the content plan (§\ref{sec:eval_planning}). In §\ref{sec:wikifluent_test}, we include an intrinsic evaluation of our modules on the \textsc{WikiFluent} test set. 

\begin{table*}[t]\centering\footnotesize
  \begin{tabular}{l p{13cm}} \toprule
      \textbf{Input}   & \textit{(Allen Forrest; background; solo singer), (Allen Forrest; genre; Pop music), (Allen Forrest; birthPlace; Dothan, Alabama)} \\
      \textbf{Templ.} &  Allen Forrest is a solo singer. Allen Forrest performs Pop music. Allen Forrest was born in Dothan, Alabama. \\
      \textbf{Model} & \lightblue{Allen Forrest is a solo singer who performs Pop music. He was born in Dothan, Alabama.} \\
      \textbf{Human} & Born in Dothan, Alabama, Allen Forrest has a background as a solo singer and was a pop artist.\Bstrut \\\hdashline[0.5pt/2pt] 
      \textbf{Input}   & \textit{name[Wildwood], eatType[restaurant], food[French], area[riverside], near[Raja Indian Cuisine]}\Tstrut \\
      \textbf{Templ.} & Wildwood is a restaurant. Wildwood serves French food. Wildwood is in the riverside. Wildwood is near Raja Indian Cuisine. \\
      \textbf{Model} & \lightblue{Wildwood is a restaurant serving French food. It is in the riverside near Raja Indian Cuisine.} \\
      \textbf{Human} & A amazing French restaurant is called the Wildwood. The restaurant is near the Raja Indian Cuisine in riverside. They love kids. \\ \bottomrule
     \end{tabular}
  \caption{Example outputs of our model (\textsc{3-stage}, filtered). See Appendix \ref{app:examples} for more examples.}\label{tab:ex1}
  \end{table*}

\subsection{Automatic Metrics}
\label{sec:eval_auto}
\begin{table}[t]
  \centering \small
  \begin{tabular}{llcccc} \toprule
   & & \textbf{B} &\textbf{M} & \textbf{O} & \textbf{H} \\\midrule

   \multicolumn{2}{l}{$\text{UPF-FORGe}^*$}    & 38.65 & 39.00 & 0.075 & 0.101 \\
   \multicolumn{2}{l}{$\textsc{Melbourne}^*$}    & 45.13 & 37.00 & 0.237 & 0.202 \\
   \multicolumn{2}{l}{$\text{\citet{ke2021jointgt}}^{\dagger *}$}  & 66.14  & 47.25 & - & - \\ 
   \multicolumn{2}{l}{$\text{\citet{laha2020scalable}}^{\dagger}$}  & 24.80  & 34.90   & - & -\Bstrut \\\hdashline[0.5pt/2pt] 
   \multicolumn{2}{l}{\baselinecopy{}}    & 37.18 & 38.77 & 0.000 & 0.000\Tstrut \\\midrule
   \multirow{3}{*}{\textit{full}} & \textsc{3-stage} & 42.92 & 39.07 & 0.051 & 0.148 \\
                                  & \textsc{2-stage} & 42.90  & 39.28 & \textbf{0.043} & 0.125 \\
                                  & \textsc{1-stage} & 39.08 & 38.94 & 0.071 & 0.204\Bstrut \\\hdashline[0.5pt/2pt] 
    \multirow{3}{*}{\textit{filtered}} & \textsc{3-stage} & 43.19 & 39.13 & 0.152 & \textbf{0.073}\Tstrut \\
                                       & \textsc{2-stage} & \textbf{43.49} & \textbf{39.32} & 0.146 & 0.096 \\
                                       & \textsc{1-stage} & 42.99 & 38.81 & 0.202 & 0.093 \\ \bottomrule
  \end{tabular}
  \caption{Automatic metrics on WebNLG. B = BLEU, M = METEOR, O = omissions / \# facts, H = hallucinations / \# examples. The systems marked with asterisk (*) are trained on the WebNLG dataset. Results for the systems marked with $\dagger$ are taken from the respective works.}
  \label{tab:webnlg}
\end{table}

\begin{table}[ht]
  \centering \small
  \begin{tabular}{ll cccc} \toprule
    & & \textbf{B} & \textbf{M} & \textbf{O} & \textbf{H}          \\\midrule
    \multicolumn{2}{l}{\textsc{TGen}$^*$}                 & 40.73 &37.76 &    0.016 & 	0.083  \\
    \multicolumn{2}{l}{\citet{harkous2020have}$^*$}\hspace{-2mm}     & 43.60 & 39.00 & - & -\Bstrut \\\hdashline[0.5pt/2pt] 
    \multicolumn{2}{l}{\baselinecopy{}}                 & 24.19	& 34.89 & 0.000 & 0.000\Tstrut \\ \midrule 
   \multirow{3}{*}{\textit{full}}  & \textsc{3-stage}    & \textbf{36.04} & 36.95 & \textbf{0.001} & \textbf{0.001}   \\
                                   & \textsc{2-stage}    & 35.84 & 36.91 &          \textbf{0.001} & \textbf{0.001}   \\
                                  & \textsc{1-stage}    & 30.81 & 36.01 & 0.009 & 0.122\Bstrut \\\hdashline[0.5pt/2pt]
    \multirow{3}{*}{\textit{filtered}} & \textsc{3-stage}& 35.88 & 36.95 &          \textbf{0.001} & \textbf{0.001}\Tstrut \\
                                       & \textsc{2-stage}& 36.01 & \textbf{36.99} & \textbf{0.001} & \textbf{0.001} \\
                                       & \textsc{1-stage}& 34.08 & 36.32 & 0.012 & 0.050  \\ \bottomrule
  \end{tabular}
  \caption{Automatic metrics on E2E. B = BLEU, M = METEOR, O = omissions / \# facts, H = hallucinations / \# examples. The systems marked with asterisk (*) are trained on the E2E dataset. The results for \citet{harkous2020have} are taken from their work.} 
  \label{tab:e2e}
\end{table}

Following prior work, we use BLEU \cite{papineni2002bleu} and METEOR \cite{banerjee2005meteor} to evaluate the outputs against the human references.\footnote{We use the implementation from \url{https://github.com/tuetschek/e2e-metrics}.} We also evaluate the number of omission and hallucination errors (i.e., facts missing or added, respectively) using a metric from \citet{duvsek2020evaluating} based on a RoBERTa model \cite{liu2019roberta} pretrained on natural language inference (NLI).\footnote{We additionally evaluated the outputs on the E2E dataset using the provided pattern-based slot error script. See Appendix \ref{app:results} for details.}

We include a diverse set of baselines for comparison. For \textbf{WebNLG} (see Table \ref{tab:webnlg}), we compare our systems with the results of:
\begin{itemize}[nosep,leftmargin=17pt]
  \item UPF-FORGe and \textsc{Melbourne} -- systems (grammar-based and supervised, respectively) from the first run of WebNLG Challenge \cite{gardent2017webnlg},
  \item  \citet{ke2021jointgt} -- a state-of-the-art system with a structure-aware encoder and task-specific pretraining,
  \item \citet{laha2020scalable} -- the only other (to our knowledge) zero-shot D2T generation system applied to WebNLG.
\end{itemize}

For \textbf{E2E} (see Table \ref{tab:e2e}), we compare our systems with the results of:
\begin{itemize}[nosep,leftmargin=17pt]
  \item \textsc{TGen} \cite{duvsek2015training} -- the baseline system for the E2E Challenge \cite{duvsek2020evaluating_challenge},
  \item \citet{harkous2020have} -- a state-of-the-art supervised system on cleaned E2E data.
\end{itemize}

For both datasets, \baselinecopy{} denotes the baseline of copying the facts without further processing.

The automatic evaluation shows that our systems consistently outperform the \baselinecopy{} baseline (e.g., $\sim$12 BLEU points for E2E), which is already strong thanks to our manually curated set of templates.\footnote{On WebNLG, our \baselinecopy{} baseline achieves 37.18 BLEU points, compared to 24.80 BLEU points of the \textit{full system} of \citet{laha2020scalable}, which uses automatic template generation.} Automatic scores also suggest that our systems are comparable with some older supervised systems. Nevertheless, our systems still underperform the state-of-the-art supervised systems. For this reason, we further focus on manual \textit{error analysis} in §\ref{sec:eval_manual} to pinpoint the current shortcomings of our approach.

The \textsc{2-stage} system is generally on par with the \textsc{3-stage} system or better, which indicates that explicit aggregation using the \textsc{agg} model may not be necessary. However, an advantage of having a separate aggregation module is the possibility to control the aggregation step explicitly. The models using the filtered version of the corpus generally produce better results, although they also bring in a larger number of omissions.

\subsection{Manual Error Analysis}

\label{sec:eval_manual}
\begin{table}[t]
  \centering\small
  \begin{tabular}{>{\hspace{-1mm}}l >{\hspace{-1mm}}l c >{\hspace{-2mm}}c >{\hspace{-2mm}}c >{\hspace{-2mm}}c >{\hspace{-2mm}}c >{\hspace{1mm}}c >{\hspace{-2mm}}c >{\hspace{-2mm}}c >{\hspace{-2mm}}c >{\hspace{-2mm}}c} \toprule
    & & \multicolumn{5}{c}{\bf WebNLG} & \multicolumn{5}{c}{\bf E2E} \\
    & & \textbf{H} & \textbf{I} & \textbf{O} & \textbf{R} & \textbf{G} & \textbf{H} & \textbf{I} & \textbf{O} & \textbf{R} & \textbf{G}        \\\midrule
   \multirow{3}{*}{\rotatebox[origin=c]{90}{\textit{full}}} 
    & \textsc{3-stage} &   3 &  39 &  2 &  2 &  16 &  0 &  1 &  0 &  0 &  17 \\
    & \textsc{2-stage} &   8 &  36 &  1 &  5 &  16 &  1 &  1 &  0 &  1 &  23 \\
    & \textsc{1-stage} &   28 &  27 &  6 &  10 &  20 &  17 &  0 &  1 &  79 &  45\Bstrut \\\hdashline[0.5pt/2pt]
   \multirow{3}{*}{\rotatebox[origin=c]{90}{\textit{filtered}}} 
    & \textsc{3-stage} &   2 &  37 &  2 &  1 &  15 &  0 &  0 &  0 &  0 &  17\Tstrut \\
    & \textsc{2-stage} &   5 &  32 &  1 &  2 &  14 &  0 &  0 &  0 &  0 &  11 \\
    & \textsc{1-stage} &   8 &  40 &  6 &  6 &  16 &  11 &  2 &  1 &  41 &  22 \\\bottomrule    
\end{tabular}\caption{Number of manually annotated errors on 100 examples: H = hallucinations, I = incorrect fact merging, O = omissions, R = redundancies, G = grammar errors or disfluencies.}
  \label{tab:manual}
  \end{table}

Since automatic performance metrics do not provide insights into specific weaknesses of the system \cite{van2021underreporting}, we manually examined 100 outputs of the models. We counted the number of errors: factual (hallucinations, omissions, incorrect fact merging, redundancies) and grammatical. The results are summarized in Table \ref{tab:manual}.

The \textsc{1-stage} model (which has to order the facts implicitly) tends to repeat the facts in the text (especially in E2E) and produces frequent hallucinations. These problems are largely eliminated with the \textsc{2-stage} and \textsc{3-stage} models, which produce almost no hallucinations or omissions. 
However, the outputs on WebNLG for all systems suffer from semantic errors resulting from merging of unrelated facts. This mostly happens with unrelated predicates connected to the same subject/object (e.g.\ “X was born in Y”, “X worked as Z” expressed as “X worked as Z in Y”; see Appendix \ref{app:examples} for examples). This behavior is the main obstacle to ensure factual consistency of the output. As a possible remedy, we propose explicitly controlling the semantics of sentence fusion \cite{ben2020semantically}, e.g. using a variant of constrained decoding \cite{balakrishnan2019constrained,wang2021mention}.

On the E2E data, which has a simpler triple structure (all predicates share the same subject), the outputs are generally consistent and the \textsc{2-stage} and \textsc{3-stage} models exhibit almost no semantic errors. Grammar errors and disfluencies stem mainly from over-eager paragraph compression or from artifacts in our templates and are relatively minor (e.g., missing “is” in “serves French food and family-friendly”).

\subsection{Content Planning}
\label{sec:eval_planning}

Following \citet{su2021plan} and \citet{zhao2020bridging}, we report the accuracy and BLEU-2 score of our \textbf{ordering model} on WebNLG against the human-generated plans from \citet{ferreira2018enriching}. The results are listed in Table \ref{tab:cp} and compared against a \textsc{random} baseline (random ordering) and prior work. The results show that although our approach again lags behind state-of-the-art supervised approaches, it can outperform both the random baseline and the Transformer-based approach from \citet{ferreira2019neural} while not using any in-domain examples.

\begin{table}[t]
  \centering\small
  \begin{tabular}{lcc} \toprule
   & \textbf{B-2} & \textbf{Acc}  \\ \midrule
   Transformer \cite{ferreira2019neural}$^\dagger$ & 52.20 & 0.35\\
   Step-by-step \cite{moryossef2019step}$^\dagger$ &70.80 & 0.47 \\
   PLANENC \cite{zhao2020bridging}$^\dagger$ & 80.10 & 0.62 \\
   Plan-then-generate \cite{su2021plan}$^\dagger$ &84.97 & 0.72\Bstrut \\\hdashline[0.5pt/2pt]
   \textsc{random} & 47.00 &	0.29\Tstrut \\ \midrule
   Ours (BART+ptr)  & 59.10 &	0.48 \\ \bottomrule
  \end{tabular}
  \caption{Evaluation of our zero-shot ordering model based on \citet{calizzano2021ordering}. B-2 = BLEU-2, Acc = accuracy. The results marked with $\dagger$ are copied from the respective papers.}
  \label{tab:cp}
  \end{table}

We also evaluate the accuracy of our \textbf{aggregation model}, using triples ordered according to the plans from \citet{ferreira2018enriching} as input. The accuracy is 0.33 per example and 0.62 per sentence boundary (random baseline is 0.23 and 0.50, respectively). The results show that although our approach is better than the random baseline, there is still room for improvement.

Finally, we manually evaluate how the \textbf{PC model} follows the content plan (i.e., keeping the predefined order and aggregating the sentences according to the delimiters) using 100 randomly chosen examples with more than 1 triple on WebNLG and E2E. We find that the model follows the content plan in 95\% and 100\% of cases, respectively. The incorrect cases include a fact not properly mentioned or an extra boundary between sentences without a separator. We can thus conclude that the pretraining task successfully teaches the PC model to follow a given content plan.

\subsection{Intrinsic Evaluation}
\label{sec:wikifluent_test}
Aside from the main D2T generation results, we also provide an intrinsic evaluation of our pipeline modules on the \mbox{\textsc{WikiFluent}} test sets. We evaluated the ordering, aggregation, and paragraph compression modules trained on the \textit{full} \textsc{WikiFluent} corpus. The results for both \textit{full} and \textit{filtered} test sets are summarized in Table \ref{tab:wikifluent_test}. The PC model achieves high scores, which follows from the fact that we provide it with ground truth content plans (i.e., the ordering and aggregation plan corresponding to the original paragraph). Accuracy of the ordering and aggregation modules is comparable to their performance on D2T datasets.

\begin{table}[t]
  \centering\small
  \begin{tabular}{ll cc} \toprule
    & &  \textbf{test (full)} & \textbf{test (filt.)}          \\\midrule
    \multirow{2}{*}{\textsc{ord}} & BLEU-2   & 64.8 &	71.9  \\
    & Accuracy     & 0.70 &	0.77   \\ \midrule
    \multirow{2}{*}{\textsc{Agg}} & Acc. per example     & 0.68  &	 0.68 \\
    & Acc. per sent. bound.     & 0.93  &	0.93   \\ \midrule
    \multirow{2}{*}{\textsc{PC}} & BLEU     & 90.72  &	91.60    \\
    & METEOR     & 63.89 &	  65.03   \\ \bottomrule
  \end{tabular}
  \caption{Result of individual pipeline modules on the \textsc{WikiFluent} test sets (full / filtered). The metrics correspond to the metrics used for evaluating the modules for D2T generation.}
  \label{tab:wikifluent_test}
\end{table}

\section{Future Work}
Our experiments outline several possible future research directions. Automatic generation of facts without using hand-crafted templates (cf. §\ref{sec:templates_model}) could allow applying zero-shot generation systems to datasets with a large number of predicates, such as ToTTo \cite{parikh2020totto}. The task of paragraph compression could be used as a task-specific pretraining \cite{gururangan2020don} for more efficient finetuning of D2T models, e.g.,~with a small amount of clean data. Consistency checks may be introduced in the pipeline to control the output from the modules, and individual modules may be improved by using more efficient model architectures.

More research is also needed regarding the main shortcoming of our approach, i.e., the semantic errors stemming from merging of facts in improper ways. As we suggested in §\ref{sec:eval_manual}, explicitly controlling the semantics of sentence fusion could help to mitigate this issue, while still keeping the advantages of a zero-shot approach.

\section{Conclusion}
\label{sec:discussion}
We presented an approach for zero-shot D2T generation. The approach uses a pipeline of PLMs trained on general-domain lexical operations over natural language. The pipeline builds upon traditional approaches and consists of three interpretable intermediate steps. By avoiding noisy human-written references from the D2T datasets, our models produce more semantically consitent output. We believe that training models for zero-shot D2T generation using large cross-domain corpora will help to build D2T generation systems with good performance across various domains.

\section{Limitations and Broader Impact}
We study zero-shot D2T generation with the focus on generating descriptions for RDF triples. Although the task of D2T generation has numerous applications, using neural models for D2T generation (especially in the zero-shot context) is still limited to experimental settings \cite{dale2020natural}. Similarly to other recent approaches for D2T generation, our approach relies on PLMs, which are known to reflect the biases in their pretraining corpus \cite{bender2021dangers,rogers-2021-changing}. Our system may therefore rely on spurious correlations for verbalizing e.g.~gender or occupation of the entities. Since we cannot guarantee the factual correctness of the outputs of our system, the outputs should be used with caution.

On the flip side, our approach helps to reduce the number of omissions and hallucinations stemming from noise in human-written references. Our work thus contributes to the general aim of D2T generation in conveying the data semantics accurately and without relying on implicit world knowledge.

\section*{Acknowledgements}

This research was supported by Charles University projects GAUK 140320, SVV 260575 and PRIMUS/19/SCI/10, an Apple NLU Research Grant for Heriot-Watt University and Charles University, and by the European Research Council (Grant agreement No. 101039303 NG-NLG). 
It used resources provided by the LINDAT/CLARIAH-CZ Research Infrastructure (Czech Ministry of Education, Youth and Sports project No. LM2018101).

\bibliography{custom}
\bibliographystyle{acl_natbib}

\appendix
\section{Data Statistics}
\label{app:stats}

Statistics for the datasets described in the paper are listed in Table \ref{tab:stats}.

\section{Templates}
\label{app:templates}
The templates for our datasets are single-sentence and mostly clear-cut verbalizations of the predicates. The templates were created by one of the authors who had only the input data at their disposal, i.e. without using human references.

We have also considered extracting the templates for WebNLG from the training data by delexicalizing single-triple examples. However, the examples are noisy and such data would not be available in a zero-shot setup, which is why we decided not to use this option.

Although the templates were mostly unambiguous, we had to opt for the most general version in certain cases (e.g., using \textit{country} $\rightarrow$ \textit{"<s> is from <o>"}, even though \textit{"<s> is a food from <o>."} would be possible in case the object is food).

Filling the templates also often results in minor disfluencies, e.g. \textit{nationality} $\rightarrow$ \textit{"<s> is from <o>"} will produce a missing definite article for \textit{<o> = "United States"} and ungrammatical sentence for \textit{<o> = "French people"}. In principle, the disfluencies may be fixed by rephrasing in the final step of the pipeline.

We provide all the templates we used in our experiments in our repository.

\section{Experimental Setup}
\label{app:setup}

We implemented the models for split-and-rephrase, aggregation, and paragraph compression in PyTorch Lightning \cite{paszke2019pytorch}, using the PyTorch \cite{falcon2019pytorch} version of the BART and RoBERTa models from the Huggingface library \cite{wolf2019huggingface}. 

We use the Adam \cite{kingmaB14} optimizer ($\beta_1=0.9, \beta_2=0.997, \varepsilon=1^{-9} $) with learning rate $2^{-5}$, linear scheduling and 0.1 warmup proportion; batches of size 8 and accumulating gradients with factor 4. We train the models for 1 epoch on a single GeForce RTX 3090 GPU with 24 GB RAM. Training times were approximately 24 hours for the ordering model and 3 hours for the aggregation and paragraph compression models.  We use greedy decoding in all our experiments.

For training the ordering model, we used the implementation from \citet{calizzano2021ordering} \footnote{\url{https://github.com/airKlizz/passage-ordering}} including their training parameters. We will integrate the ordering model into our framework.

\section{Additional Results}

\label{app:results}

We provide evaluation of semantic accuracy on the E2E dataset as evaluated with the slot-error script based on matching regular expressions in Table \ref{tab:e2e_extra}.\footnote{\url{https://github.com/tuetschek/e2e-cleaning/blob/master/slot_error.py}}

\begin{table}[ht]
  \centering\small
  \begin{tabular}{ll ccc} \toprule
    &  & \textbf{miss} & \textbf{add} & \textbf{miss+add}          \\\midrule
    \multicolumn{2}{l}{\textsc{TGen}}                 & 0.0060 &	0.0433 &	0.0016  \\
    \multicolumn{2}{l}{\baselinecopy{}}                 & 0.0000 &  0.0000 & 0.0000   \\ \midrule 
   \multirow{3}{*}{\textit{full}}  & \textsc{3-stage}    &0.0238	& 0.0698 &	0.0060 \\
                                   & \textsc{2-stage}    &0.0054	& 0.0363 &	0.0000 \\
                                  & \textsc{1-stage}     &0.0043	& 0.0330 &	0.0000\Bstrut \\\hdashline[0.5pt/2pt]
    \multirow{3}{*}{\textit{filtered}} & \textsc{3-stage}&0.0444	& 0.0487 &	0.0076\Tstrut \\
                                       & \textsc{2-stage}&0.0043	& 0.0368 &	0.0000 \\
                                       & \textsc{1-stage}&0.0043	& 0.0347 &	0.0000 \\ \bottomrule
  \end{tabular}
  \caption{Proportion of output examples with missed only, added only, and both missed and added facts, according to the regex-based E2E slot error script.}
  \label{tab:e2e_extra}
\end{table}

However, please note that our manual investigation of a sample of the data shows that the majority of the errors identified in our model outputs are false. For example, the following regular expression used in the slot-error script:
$$
\texttt{prices?(?: range)?(?: \\w+){0,3} high}
$$
\noindent matches \textit{"(...) price range and high customer rating (...)"}, incorrectly classifying the presence of the extra slot \textit{priceRange[high]}. This problem is magnified by the consistent outputs of our models, which tend to repeat certain patterns. However, we also manually identified several cases in which an error was found correctly, e.g.\ the model hallucinating \textit{"3 out of 4 customer rating"} instead of \textit{"3 out of 5 customer rating"}.

\section{Example Outputs}

\label{app:examples}
Tables \ref{tab:ex2}, \ref{tab:ex3}, \ref{tab:ex4}, and \ref{tab:ex5} show examples of behavior of our models on the \textbf{WebNLG dataset}. Tables \ref{tab:ex6} and \ref{tab:ex7} show examples of behavior of our models on the \textbf{E2E dataset}.

The \green{green} color marks the model outputs which are completely correct, the \red{red} color marks the errors. For better readability of the input format, we add numeric order identifiers for the individual facts (bold, in squared brackets). These are subsequently used as references in the Order and Aggregation rows. Note that zero-th input in E2E is used as a subject in the RDF triples.

\begin{table*}[htbp]
  \centering 
  \begin{tabular}{l cccccccc}\toprule
      & \bf  \#train & \bf \#dev & \bf \#test & \bf tok/src & \bf tok/tgt & \bf sent/src & \bf sent/tgt & \bf \#temp. \\  \midrule
    WebNLG & 18,102 & 870 & 1,862 & 26.8 & 22.6 & 3.0 & 1.4 & 354 \\
    Clean E2E  &  33,236 & 4,299 & 1,847 & 29.2 & 22.3 & 4.2 & 1.5  & 8 \\ \midrule
    \textsc{WikiFluent}-\textit{full}  & 915,855 & 9,346 & 9,346 &52.9 & 41.1 & 3.9 & 2.0 & -  \\
    \textsc{WikiFluent}-\textit{filtered}  & 700,517 & 7,149 & 7,149 & 45.6 & 35.4 & 3.4 & 1.8 & - \\ \bottomrule

  \end{tabular}
  \caption{Number of examples (train / dev / test), average number of tokens per source and target, average number of sentences per source and target (after filling the templates for the D2T datasets), total number of templates.}
  \label{tab:stats}
\end{table*}

\begin{table*}[htbp]\centering
  \begin{tabular}{l p{13cm}} \toprule
      \textbf{Input}   &  \textit{\textbf{[1]} (Andrews County Airport; elevationAboveTheSeaLevel (in metres); 973.0) \newline \textbf{[2]} (Andrews County Airport; runwayLength; 896.0) \newline \textbf{[3]} (Andrews County Airport; location; Texas)} \\
      \textbf{Facts} & Andrews County Airport is 973.0 metres above the sea level. \newline Andrews County Airport runway is 896.0 m long. \newline Andrews County Airport is located in Texas. \\
      \textbf{Order} & 3 2 1 \\
      \textbf{Aggregation} & 3 <sep> 2 1 \\
      \textbf{Models (all)} & \green{Andrews County Airport is located in Andrews County, Texas. Its runway is 896.0 m long and 973.0 metres above the sea level.} \\
      \textbf{Human} & The runway length of Andrews County Airport (located in Texas and 973.0 above sea level) is 896. \\ \bottomrule
     \end{tabular}
  \caption{Correct behavior of the models on WebNLG. All the models \textsc{(\{1,2,3\}-stage} \textit{full/filtered}) produce the same output.}\label{tab:ex2}
  \end{table*}

  \begin{table*}[htpb]\centering
    \begin{tabular}{l p{13cm}} \toprule
        \textbf{Input}   & \textit{\textbf{[1]} (Akron, Ohio; country; United States) \newline \textbf{[2]} (United States; language; English language)} \\
        \textbf{Facts} & Akron, Ohio is from United States.\newline English language is spoken in United States. \\
        \textbf{Order} & 1 2 \\
        \textbf{Aggregation} & 1 <sep> 2 \\
        \textbf{3-stage full} &Akron, Ohio is from United States. English language is spoken in  \red{Akron, Ohio,} United States\red{, which is located in the U.S. state of Ohio.} \\
        \textbf{2-stage full} &Akron, Ohio is from United States. English language is spoken in \red{Akron, Ohio.} \\
        \textbf{1-stage full} &English language is spoken in the United States. Akron, Ohio is from the state of Ohio \red{in the U.S. state of Illinois.}\\
        \textbf{3-stage filtered} &\green{Akron, Ohio is from United States. English language is spoken in the United States of America.}\\
        \textbf{2-stage filtered} &Akron, Ohio is from United States. English language is spoken in \red{Akron, Ohio.} \\
        \textbf{1-stage filtered} &English language is spoken in the United States. Akron, Ohio is from the \red{United Kingdom.}\\
        \textbf{Human} & The country of Akron, Ohio, is United States, where English is the language spoken. \\ \bottomrule
       \end{tabular}
    \caption{Incorrect behavior on WebNLG: besides the minor disfluencies caused by the templates (\textit{"Akron, Ohio is from..."}), the models (except for \textsc{3-stage} \textit{filtered}) tend to hallucinate and merge the facts incorrectly.}
    \label{tab:ex3}
    \end{table*}

    \begin{table*}[hptb]\centering
      \begin{tabular}{l p{13cm}} \toprule
          \textbf{Input}   & \textit{\textbf{[1]} (Juan Perón; party; Labour Party (Argentina)) \newline \textbf{[2]} (Alberto Teisaire; inOfficeWhilePresident; Juan Perón) \newline \textbf{[3]} (Alberto Teisaire; nationality; Argentina) \newline \textbf{[4]} (Argentina; language; Spanish language)} \\
          \textbf{Facts} & Juan Perón belongs to the Labour Party Argentina. \newline Alberto Teisaire was in office while Juan Perón was a president. \newline Alberto Teisaire is from Argentina. \newline Spanish language is spoken in Argentina. \\
          \textbf{Order} & 3 2 1 4  \\
          \textbf{Aggregation} & 3 2 <sep> 1 <sep> 4 \\
          \textbf{3-stage full} & Alberto Teisaire is from Argentina, \red{who was in office} while Juan Perón was a president. \red{He} belongs to the Labour Party Argentina. Spanish language is spoken in Argentina. \\
          \textbf{2-stage full} & Alberto Teisaire is from Argentina. He was in office while Juan Perón was a president. \red{He} belongs to the Labour Party Argentina. Spanish language is spoken in Argentina. \\
          \textbf{1-stage full} &\green{Alberto Teisaire is from Argentina. He was in office while Juan Perón was a president. Spanish language is spoken in Argentina. Perón belongs to the Labour Party Argentina.}\\
          \textbf{3-stage filtered} & Alberto Teisaire is from Argentina, in office while Juan Perón was a president. \red{He} belongs to the Labour Party Argentina. Spanish language is spoken in Argentina. \\
          \textbf{2-stage filtered} & Alberto Teisaire is from Argentina. He was in office while Juan Perón was a president. \red{He} belongs to the Labour Party Argentina. Spanish language is spoken in Argentina. \\
          \textbf{1-stage filtered} & \green{Alberto Teisaire is from Argentina. He was in office while Juan Perón was a president. Spanish language is spoken in Argentina. Juan belongs to the Labour Party Argentina.}\\
          \textbf{Human} & Alberto Teisaire, is a national of Argentina, where Spanish is spoken. Alberto Teisaire, was in office while Juan Perón was president. Juan Perón was a member of the Labour Party in Argentina. \\ \bottomrule
         \end{tabular}
      \caption{Incorrect behavior on WebNLG: by using a pronoun "He", the models tend to imply that \textit{Alberto Teisaire} (instead of \textit{Juan Perón)} belongs to the Labour Party.}
      \label{tab:ex4}
      \end{table*}

      \begin{table*}[hptb]\centering
        \begin{tabular}{l p{13cm}} \toprule
            \textbf{Input}   &  \textit{\textbf{[1]} (Alfa Romeo 164; relatedMeanOfTransportation; Fiat Croma) \newline \textbf{[2]} (Alfa Romeo 164; assembly; Italy) \newline \textbf{[3]} (Italy; capital; Rome)} \\
            \textbf{Facts} & Alfa Romeo 164 is related to Fiat Croma. \newline Alfa Romeo 164 was assembled in Italy. \newline Italy's capital is Rome. \\
            \textbf{Order} & 2 3 1 \\
            \textbf{Aggregation} & 2 3 <sep> 1 \\
            \textbf{3-stage full} & Alfa Romeo 164 was assembled \red{in Italy's capital, Rome}. It is related to Fiat Croma. \\
            \textbf{2-stage full} &Alfa Romeo 164 was assembled \red{in Italy's capital, Rome}. It is related to Fiat Croma. \\
            \textbf{1-stage full} & Alfa Romeo 164 was assembled in Italy \orange{and related} to Fiat Croma. Italy's capital is Rome.\\
            \textbf{3-stage filtered} & Alfa Romeo 164 was assembled \red{in Italy's capital, Rome.} It is related to Fiat Croma.\\
            \textbf{2-stage filtered} &Alfa Romeo 164 was assembled \red{in Italy's capital, Rome.} It is related to Fiat Croma. \\
            \textbf{1-stage filtered} &Alfa Romeo 164 was assembled in Italy. It is related to Fiat Croma. \red{It was assembled at the Italian capital is Rome.} \\
            \textbf{Human} & The Fiat Croma and the Alfa Romeo 164 are both cars. The latter was built in Italy where the capital is Rome. \\ \bottomrule
           \end{tabular}
        \caption{Incorrect behavior on WebNLG: the models tend to merge the country of vehicle's assembly and its capital.}\label{tab:ex5}
        \end{table*}

      \begin{table*}[htpb]\centering
        \begin{tabular}{l p{13cm}} \toprule
            \textbf{Input}   &  
            \textit{\textbf{[0]} name[The Cricketers]
            \newline \textbf{[1]} eatType[restaurant]
            \newline \textbf{[2]} near[All Bar One]
            \newline \textbf{[3]} priceRange[cheap]
            \newline \textbf{[4]} food[Chinese]
            \newline \textbf{[5]} customerRating[average]
            \newline \textbf{[6]} familyFriendly[yes]} \\
            \textbf{Facts} & 
            The Cricketers is a restaurant. 
            \newline The Cricketers is near All Bar One. 
            \newline The Cricketers has cheap price range. 
            \newline The Cricketers serves Chinese food. 
            \newline The Cricketers has average customer rating. 
            \newline The Cricketers is family-friendly. 
            \\
            \textbf{Order} & 1 4 2 6 3 5 \\
            \textbf{Aggregation} & 1 4 2 <sep> 6 3 5 \\
            \textbf{3-stage full} & \green{The Cricketers is a restaurant serving Chinese food near All Bar One. It is family-friendly, has cheap price range and average customer rating.} \\
            \textbf{2-stage full} & \green{The Cricketers is a restaurant serving Chinese food near All Bar One. It is family-friendly with cheap price range and average customer rating.} \\
            \textbf{1-stage full} & \green{The Cricketers is a restaurant serving Chinese food. It is family-friendly and has cheap price range. It serves Chinese food near All Bar One. It has average customer rating.}\\
            \textbf{3-stage filtered} & \green{The Cricketers is a restaurant serving Chinese food near All Bar One. It is family-friendly with cheap price range and average customer rating.}\\
            \textbf{2-stage filtered} & \green{The Cricketers is a restaurant serving Chinese food near All Bar One. It is family-friendly with cheap price range and average customer rating.} \\
            \textbf{1-stage filtered} & \green{The Cricketers is a restaurant serving Chinese food. It is family-friendly and has average customer rating. It serves Chinese food near All Bar One. It has cheap price range}.\\
            \textbf{Human} & For a central family friendly venue serving cheap Chinese food, try The Cricketers restaurant. It has average customer ratings and is near All Bar One. \\ \bottomrule
           \end{tabular}
        \caption{Correct behavior of the models on E2E: the models can deal even with a higher amount of input values.}\label{tab:ex6}
        \end{table*}

        \begin{table*}[htbp]\centering
          \begin{tabular}{l p{13cm}} \toprule
              \textbf{Input}   &  
              \textit{\textbf{[0]} name[The Phoenix] 
              \newline \textbf{[1]} area[riverside]
              \newline \textbf{[2]} priceRange[cheap]
              \newline \textbf{[3]} familyFriendly[yes]
              \newline \textbf{[4]} near[Raja Indian Cuisine]
              \newline \textbf{[5]} eatType[restaurant]} \\
              \textbf{Facts} & 
              The Phoenix is in the riverside.
              \newline The Phoenix has cheap price range.
              \newline The Phoenix is family-friendly.
              \newline The Phoenix is near Raja Indian Cuisine.
              \newline The Phoenix is a restaurant.
              \\
              \textbf{Order} & 5 1 4 <sep> 3 2 \\
              \textbf{Aggregation} & 5 1 4 3 2 \\
              \textbf{3-stage full} &\green{The Phoenix is a restaurant in the riverside near Raja Indian Cuisine. It is family-friendly with cheap price range.} \\
              \textbf{2-stage full} & \green{The Phoenix is a restaurant in the riverside near Raja Indian Cuisine. It is family-friendly with cheap price range.} \\
              \textbf{1-stage full} & The Phoenix is a restaurant near Raja Indian Cuisine in the riverside. It is family-friendly and has cheap price range. \red{It has cheap food range.} \\
              \textbf{3-stage filtered} & \green{The Phoenix is a restaurant in the riverside near Raja Indian Cuisine. It is family-friendly with cheap price range.} \\
              \textbf{2-stage filtered} & \green{The Phoenix is a restaurant in the riverside near Raja Indian Cuisine. It is family-friendly with cheap price range. }\\
              \textbf{1-stage filtered} &  The Phoenix is a restaurant near Raja Indian Cuisine in the riverside. It is family-friendly and has cheap price range. \red{It has cheap food.} \\
              \textbf{Human} & Cheap food and a family friendly atmosphere at The Phoenix restaurant. Situated riverside near the Raja Indian Cuisine. \\ \bottomrule
             \end{tabular}
          \caption{Incorrect behavior on E2E: the \textsc{1-stage} models add redundant information to the output.}\label{tab:ex7}
          \end{table*}

\end{document}